\documentclass{article}

\usepackage{microtype}
\usepackage{graphicx}
\usepackage{subfigure}
\usepackage{booktabs} % for professional tables

\usepackage{hyperref}

\usepackage[accepted]{icml2019}
\begin{document}

\twocolumn[
\icmltitle{Visualizing Uncertainty and Saliency Maps of Deep \\ Convolutional Neural Networks for Medical Imaging Applications}

\icmlsetsymbol{equal}{*}

\begin{icmlauthorlist}
\icmlauthor{Jae Duk Seo}{to}

\end{icmlauthorlist}

\icmlaffiliation{to}
{Department of Computer Science, Ryerson University, Toronto Ontario, Canada}
\icmlcorrespondingauthor{Jae Duk Seo}{jae.duk.seo@ryerson.ca}

% You may provide any keywords that you
% find helpful for describing your paper; these are used to populate
% the "keywords" metadata in the PDF but will not be shown in the document
\icmlkeywords{Medical Imaging, Bio-medical Imaging, Machine Learning, Artificial Intelligence, Bio}
\vskip 0.3in
]

% this must go after the closing bracket ] following \twocolumn[ ...

% This command actually creates the footnote in the first column
% listing the affiliations and the copyright notice.
% The command takes one argument, which is text to display at the start of the footnote.
% The \icmlEqualContribution command is standard text for equal contribution.
% Remove it (just {}) if you do not need this facility.

\printAffiliationsAndNotice{}  
% ============================================
\begin{abstract}
Deep learning models are now used in many different industries, while in certain domains safety is not a critical issue in the medical field it is a huge concern. Not only, we want the models to generalize well but we also want to know the model's confidence respect to it's decision and which features mattered the most. Our team aims to develop a full pipeline in which not only displays the uncertainty of the model's decision but also the saliency map to show which sets of pixels of the input image contribute most to the predictions.
\end{abstract}
% ============================================

% ============================================
\section{Introduction}
\label{Introduction}
The use of machine learning models in bio-medical imaging has ever been growing since the break through of Image-net challenge in 2014. However, interpret-ability of these models are still a huge problem for industry adoption. Recently developed gradient method such as \cite{simonyan2013deep,springenberg2014striving,sundararajan2017axiomatic}, reveals which part of the image the network has focused on to make either segmentation or classification decision. While \cite{Gal} have created a theoretical framework for justifying the use of drop-out as uncertainty estimation. 

\begin{figure}[!ht]
\begin{center}
\vskip 0.1in
\centerline{\includegraphics[width=\columnwidth]{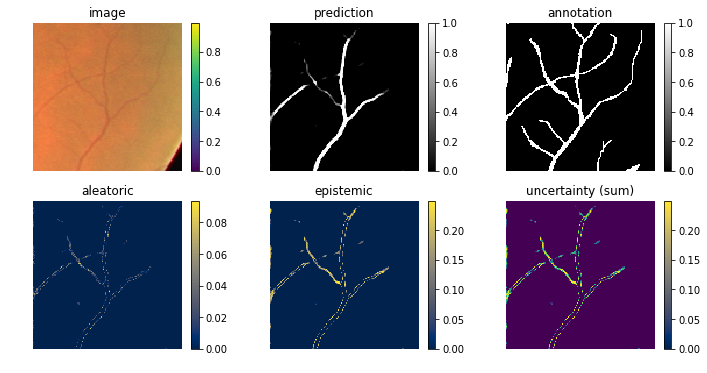}}
\caption{Segmentation result for Retinal Images along with the models Epistemic and Aleatoric uncertainty}
\label{icml-historical}
\vskip -0.6in
\end{center}
\end{figure}
% ============================================

% ============================================
\section{Problem Description}
Statistical models, such as deep neural networks, that are used in medical imaging domain must be transparent and interpretable. One way of achieving this is by outputting the model's confidence with it's decision as well as visualize which part of the original image did the model take into account the most. To best of our knowledge, this is the first attempt to predict model's uncertainty while producing it's saliency map. This is a problem since, the current standard for measuring performance of these models is classification or segmentation accuracy. While those measurement's are important, it is not the full picture. To insure trust from the patient and the medical professional, expert systems need to report their inner working mechanisms. 
% ============================================

% ============================================
\section{Design Solution}
Thanks to recent theoretical development done related to Bayesian neural networks now it is possible to produce both epistemic and aleatoric uncertainty. Additionally, methods that reveals which input features contributed the most for network's decision have been developed, as early as back in 2014. We aim to combine both of these procedures as a small step towards creating transparent machine learning models for medical imaging. 

\subsection{Bayesian Deep Learning}
There are two different types of uncertainty, epistemic uncertainty which captures the confidence within the used statistical model. This uncertainty can be explained away given enough data, and is often referred to as model uncertainty. And aleatoric uncertainty captures the uncertainty with respect to information which our data cannot explain. And based on the theoretical development of a methodology called 'Dropout as a Bayesian Approximation' both can be measured using drop-out. 

\subsection{Saliency Mapping}
For both classification and segmentation results knowing which part of the input data is mostly important to the model is extremely beneficial. Especially for medical diagnosis, the medical professional can actually see if the model is focusing on the right places. Here we apply the gradient methods such as integral gradients, to visualize which part of the input matters the most. 
% ============================================

% ============================================
\begin{figure}[ht]
\vskip 0.2in
\begin{center}
\centerline{\includegraphics[width=\columnwidth]{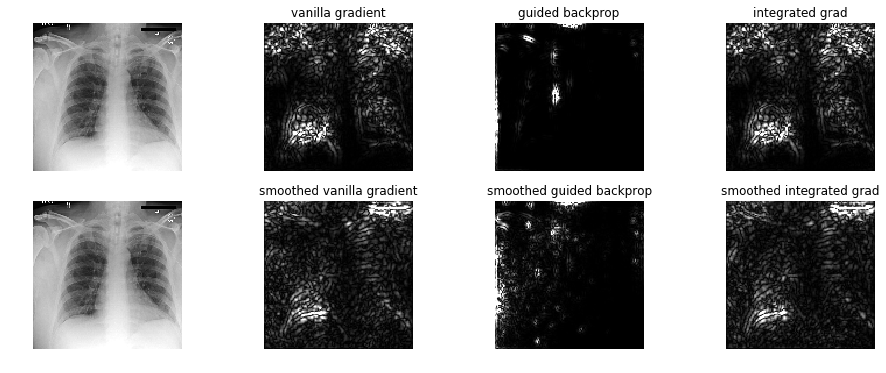}}
\caption{Different Saliency mapping visualized for chest X-Ray classification}
\label{icml-historical}
\end{center}
\vskip -0.2in
\end{figure}

\section{Application}
\subsection{Segmentation of Blood Vessels in Retinal Images}
Using the data from \cite{staal2004ridge} we trained a nine layer Unet\cite{ronneberger2015u} to perform segmentation. Which the objective function was set as the Dice coefficient loss. The model was able to achieve 0.22 loss on training set and 0.26 on validation data, indicating a high generalization power of the model. And as depicted in figure 1 by using the Monte Carlo Dropout Sampling method we can output the model's uncertainty. We can notice when the model is uncertain about the region of the blood vessel, it tends to produce very thin segmentation maps. 

\subsection{Classification on Chest X-ray images}
Next we used a pre-trained VGG 16 network\cite{Simonyan15}, and only trained a top layer to perform classification on Chest X-ray \cite{Wang_2017_CVPR} data set. And as seen in figure 2, we can observe the location where the model focuses on for classification. And each of the saliency mapping methods (and it's smoother version) gives consistent results, indicating that the left bottom region of the image is a critical region. 
% ============================================

% ============================================
\section{Implementation}
Due to restrictions related to medical imaging data, it is not realistic or feasible to expect multiple health institution to share their data. However, thanks to increase in the demand of data scientist in many health organization it is reasonable to expect that each of those organization would be interested in applying a deep learning models as possible solutions. Hence we see the most optimal approach is to create a step by step tutorial series, either in a video format or a single web-page format, so that different organizations can implement this solution in their current system. Additionally, we are aiming to expand the example use-case of these methods, specifically, we are planning to use neural networks to achieve dual task on microscopic cell images. Not only we can perform segmentation of cell regions but also count the number of cells in a particular image, while outputting uncertainty for each measurements. Finally, possible future works includes, creating a proof of concept web application where medical professionals can upload their own data and see the promising result of this methodology. 
% ============================================

% ============================================
\begin{footnotesize}
\section*{Acknowledgements}
This research was supported by Ryerson Vision lab, and the author of this paper wishes to express appreciations toward Dr. Neil Bruce and Dr. Kosta Derpanis for providing a comfortable lab space. 

\bibliography{example_paper}

\begin{thebibliography}{8}
\providecommand{\natexlab}[1]{#1}
\providecommand{\url}[1]{\texttt{#1}}
\expandafter\ifx\csname urlstyle\endcsname\relax
  \providecommand{\doi}[1]{doi: #1}\else
  \providecommand{\doi}{doi: \begingroup \urlstyle{rm}\Url}\fi

\bibitem[Gal \& Ghahramani(2016)Gal and Ghahramani]{Gal}
Gal, Y. and Ghahramani, Z.
\newblock Dropout as a bayesian approximation: Representing model uncertainty
  in deep learning.
\newblock In \emph{Proceedings of the 33rd International Conference on
  International Conference on Machine Learning - Volume 48}, ICML'16, pp.\
  1050--1059. JMLR.org, 2016.
\newblock URL \url{http://dl.acm.org/citation.cfm?id=3045390.3045502}.

\bibitem[Ronneberger et~al.(2015)Ronneberger, Fischer, and
  Brox]{ronneberger2015u}
Ronneberger, O., Fischer, P., and Brox, T.
\newblock U-net: Convolutional networks for biomedical image segmentation.
\newblock In \emph{International Conference on Medical image computing and
  computer-assisted intervention}, pp.\  234--241. Springer, 2015.

\bibitem[Simonyan \& Zisserman(2015)Simonyan and Zisserman]{Simonyan15}
Simonyan, K. and Zisserman, A.
\newblock Very deep convolutional networks for large-scale image recognition.
\newblock In \emph{International Conference on Learning Representations}, 2015.

\bibitem[Simonyan et~al.(2013)Simonyan, Vedaldi, and
  Zisserman]{simonyan2013deep}
Simonyan, K., Vedaldi, A., and Zisserman, A.
\newblock Deep inside convolutional networks: Visualising image classification
  models and saliency maps.
\newblock \emph{arXiv preprint arXiv:1312.6034}, 2013.

\bibitem[Springenberg et~al.(2014)Springenberg, Dosovitskiy, Brox, and
  Riedmiller]{springenberg2014striving}
Springenberg, J.~T., Dosovitskiy, A., Brox, T., and Riedmiller, M.
\newblock Striving for simplicity: The all convolutional net.
\newblock \emph{arXiv preprint arXiv:1412.6806}, 2014.

\bibitem[Staal et~al.(2004)Staal, Abr{\`a}moff, Niemeijer, Viergever, and
  Van~Ginneken]{staal2004ridge}
Staal, J., Abr{\`a}moff, M.~D., Niemeijer, M., Viergever, M.~A., and
  Van~Ginneken, B.
\newblock Ridge-based vessel segmentation in color images of the retina.
\newblock \emph{IEEE transactions on medical imaging}, 23\penalty0
  (4):\penalty0 501--509, 2004.

\bibitem[Sundararajan et~al.(2017)Sundararajan, Taly, and
  Yan]{sundararajan2017axiomatic}
Sundararajan, M., Taly, A., and Yan, Q.
\newblock Axiomatic attribution for deep networks.
\newblock In \emph{Proceedings of the 34th International Conference on Machine
  Learning-Volume 70}, pp.\  3319--3328. JMLR. org, 2017.

\bibitem[Wang et~al.(2017)Wang, Peng, Lu, Lu, Bagheri, and
  Summers]{Wang_2017_CVPR}
Wang, X., Peng, Y., Lu, L., Lu, Z., Bagheri, M., and Summers, R.~M.
\newblock Chestx-ray8: Hospital-scale chest x-ray database and benchmarks on
  weakly-supervised classification and localization of common thorax diseases.
\newblock In \emph{The IEEE Conference on Computer Vision and Pattern
  Recognition (CVPR)}, July 2017.

\end{thebibliography}
\bibliographystyle{icml2019}
\end{footnotesize}

\end{document}